\documentclass[sigconf]{acmart}

\usepackage{booktabs} 
 
\usepackage{wrapfig}
\usepackage{balance}
\usepackage{soul}
\usepackage{breqn}

\setcopyright{rightsretained}

\copyrightyear{2018} 
\acmYear{2018} 
\setcopyright{othergov}
\acmConference[GECCO '18]{Genetic and Evolutionary Computation Conference}{July 15--19, 2018}{Kyoto, Japan}
\acmBooktitle{GECCO '18: Genetic and Evolutionary Computation Conference, July 15--19, 2018, Kyoto, Japan}
\acmPrice{15.00}
\acmDOI{10.1145/3205455.3205646}
\acmISBN{978-1-4503-5618-3/18/07}

\begin{document}
\title{How swarm size during evolution impacts the behavior, generalizability, and brain complexity of animats performing a spatial navigation task}

\author{Dominik Fischer}
\affiliation{%
  \institution{Technical University Munich}
  \streetaddress{Arcsisstrasse 21}
  \city{Munich} 
  \country{Germany} 
  \postcode{80333}
}
\email{d.fischer@tum.de}

\author{Sanaz Mostaghim}
\affiliation{%
  \institution{University of Magdeburg}
  \city{Magdeburg} 
  \country{Germany}}
\email{sanaz.mostaghim@ovgu.de}

\author{Larissa Albantakis}
\affiliation{%
  \institution{University of Madison-Wisconsin}
  \city{Madison} 
  \state{Wisconsin} 
  \postcode{43017-6221}
}
\email{albantakis@wisc.edu}
\renewcommand{\shortauthors}{D. Fischer, S. Mostaghim, L. Albantakis}

\begin{abstract}

While it is relatively easy to imitate and evolve natural swarm behavior in simulations, less is known about the social characteristics of simulated, evolved swarms, such as the optimal (evolutionary) group size, why individuals in a swarm perform certain actions, and how behavior would change in swarms of different sizes. To address these questions, we used a genetic algorithm to evolve animats equipped with Markov Brains in a spatial navigation task that facilitates swarm behavior. The animats' goal was to frequently cross between two rooms without colliding with other animats. Animats were evolved in swarms of various sizes. We then evaluated the task performance and social behavior of the final generation from each evolution when placed with swarms of different sizes in order to evaluate their generalizability across conditions. According to our experiments, we find that swarm size during evolution matters: animats evolved in a balanced swarm developed more flexible behavior, higher fitness across conditions, and, in addition, higher brain complexity.

\end{abstract}

%
%



\maketitle
\section{Introduction}
When watching swarms in real life people often assume a global intelligence behind the swarm behavior, e.g. a flying mock of birds may seem to behave like a single organism \cite{Garnier2007}. However, we now know that individuals in a swarm often act based on local rules to achieve global goals \cite{Reynolds1987FlocksHA}. This principle underlies the development of dedicated algorithms to solve single and multi-objective optimization problems, like \textit{Particle Swarm Optimization (PSO)} or \textit{Ant Colony Optimization (ACO)}. Again, from the outside perspective, the abstract and virtual organisms seem to have swarm behavior, but their most basic rules are predefined by the optimization algorithms, e.g. in ACO all actions are predefined by the algorithm and its parameters \cite{Dorigo1996}. The observed complexity of swarm intelligence is then a result of the optimization process through interaction with the environment \cite{Ilie2013}. Using machine learning approaches, it is also possible to evolve such swarm behavior without the need of a predefined algorithm that controls the organism, which means that the hard-wired decision rules of the earlier mentioned algorithms are now replaced by unsupervised learning techniques \cite{Stanley2005a,Olson2013,Konig2009}.  

Being able to evolve swarm behavior brings up new questions, e.g. about the effects of swarm size on evolution and how swarm size during evolution influences the organisms' decision rules. While in the scope of biology there are several studies on group size effects in swarms and optimal group size for different species \cite{Pacala1996,Brown1982}, it is hardly feasible to conduct studies spanning evolutionary time-scales. Here, computational approaches using \textit{Evolutionary Algorithms (EA)} resembling evolution in nature provide new tools to address these questions. However, since the cognitive decision rules of adaptive artificial organisms are not hard-wired, but evolved over time they are readily observable but often hard to interpret.

In this study, we want to advance on these open questions by evolving animats\footnote{An animat is an artificial animal with the ability to have specific motor reactions to sensory signals and to have internal states \cite{Wilson1985}.} with swarm behavior and analyzing their internal `brain' states and decisions. We hypothesized that swarm size could have a great impact on the evolution and social interactions of the animats. For this reason, we designed a virtual experiment to test the effect of swarm size on the animats' evolution and, moreover, assess the generalizability of their evolved behavior when performing in swarms of different sizes. In particular, we simulated and evolved groups of animats equipped with \textit{Markov Brains (MB)} \cite{Hintze2017} in a novel spatial-navigation task environment that facilitated swarm behavior. As observed in previous work \cite{Dorigo2004b,Trianni2003}, we found that the final environmental task fitness during evolution depended negatively on swarm size: task difficulty increased with swarm size as the 2-dimensional task environment became more crowded. In addition, however, the evolved animats showed significant differences ($p < 0.05$ for 9 out of 10 tests) in generalizability regarding their fitness when placed in different-sized swarms, which peaked for animats evolved in swarms of medium size. Interestingly, animats evolved in swarms of medium sizes also evolved more complex, integrated brain structures, which was evaluated by measuring the largest strongly connected network component in their MBs.


\section{Related Work}
Adaptive animats equipped with MBs were first introduced by Edlund et al. \cite {edmund2011}. In several works, Olson et al. used these animats to investigate the evolution of swarm behavior in a predator-prey (co-)evolution environment \cite{Olson2016, Olson2013a, Olson2013}. By contrast, swarm behavior in the present study emerged directly as a result of the implemented selection rule (see below). The cognitive setup of animats used here to study group evolution and behavior most closely resembled the MBs described by Marstaller et al. \cite{Marstaller2013}, which could move left or right and were evolved to solve a temporal-spatial integration task. The same type of task and animats were also used later by Albantakis et al. \cite{Albantakis2014}, who evolved single animats in environments that required different degrees of context-dependent behavior and memory to investigate the evolution of integrated information \cite{Oizumi2014}, a measure of brain complexity developed with the objective to capture the quality and quantity of consciousness in organisms. 


A different approach to the evolution of artificial swarm behavior is the \textit{neuroevolving robotic operatives (NERO)} video game combined with \textit{real-time neuroevolution of augmenting topologies (rtNEAT)} \cite{Stanley2005a, Miikkulainen2012, Karpov2015}. Similar to Olson et al. \cite{Olson2016}, they also evolved swarm behavior in a predator-prey scenario, but with a learning technology based on \textit{artificial neural networks (ANN)}. Miikkulainen et al. \cite{Miikkulainen2012} reviewed the work around neuroevolution and discussed future research topics in this field. They concluded that cooperative multi-agent systems are the next frontier of neuroevolution and that research in this field is still in an early stage. Another alternative to animats equipped with MBs, could be  \textit{Intelligent Distribution Agents (IDA)} \cite{Franklin1998} or, if adding the self-learning component, \textit{Learning IDA (LIDA)} \cite{Franklin2006, Franklin2012}. The goal using this architecture was to develop a model of human cognition to investigate answers about the human brain and to apply the agents in real-life communications with humans.  

Hamann \cite{Hamann2014} showed that swarm behavior in general and the type of swarm behavior in particular is dependent on the density of the swarm.  In his work he used a one dimensional changing environment to influence the evolution. The fitness values are not comparable, since he rated the predictability of future states, not the behavior. Dorigo et al. implemented the evolution of self-organizing swarmbots and also investigated changing group size \cite{Dorigo2004b}. They showed that it is easier for smaller groups to organize themselves as for larger groups. An earlier work considered different numbers of agents in the environment\cite{Trianni2003} and also demonstrated that fitness decreases with increasing group size in a similar task of self-organization. 

Apart from the above research on artificial systems, several studies investigated the subject of different swarm sizes in biological systems. Pacala et al. \cite{Pacala1996} argue, on the example of ants foraging on food sources or nest maintenance, that a variation in swarm size implies that the organisms transfer different information and perform different tasks. It is mentioned that larger swarms can be more efficient than smaller ones, but very large swarm sizes can also be of disadvantage. In general, swarm behavior is the result of individual interaction with the environment combined with social interaction. Earlier, Brown \cite{Brown1982} presented work on the threshold of sociality, the willingness of an organism to join a swarm depending on the environmental qualities and swarm density. Here, optimal swarm size is expressed as a compromise between advantages gained by sharing costs and disadvantages arising from the faster loss of resources. 

\section{Methods}

In order to design an environment in which the animats are able to evolve swarm behavior and allow for efficient analysis of their behavior and internal states, we have identified the following constraints to frame our model: (1) Animats must be able to co-exist (multiple organisms in one environment). (2) Respecting other animats (non-egoistic behavior) should help to gain higher fitness. (3) The task should be simple enough to be solved by animats with only a small number of sensors, motors and hidden nodes. 

In this section, we describe the three main simulation components: (1) the animat design, (2) the 2-dimensional, grid-based environment design, and (3) the EA's fitness function. The EA was configured using the MABE (Modular Agent-Based Evolver) framework\footnote{\url{https://github.com/Hintzelab/MABE/}}\cite{CliffordBohmNitashCG2017} for digital evolution. If not specified otherwise, we used MABE's default parameters throughout, which can be reviewed in supplementary material A.1.

\subsection{Animat Design}

\begin{figure}[!htp]
\includegraphics[width=0.25\textwidth]{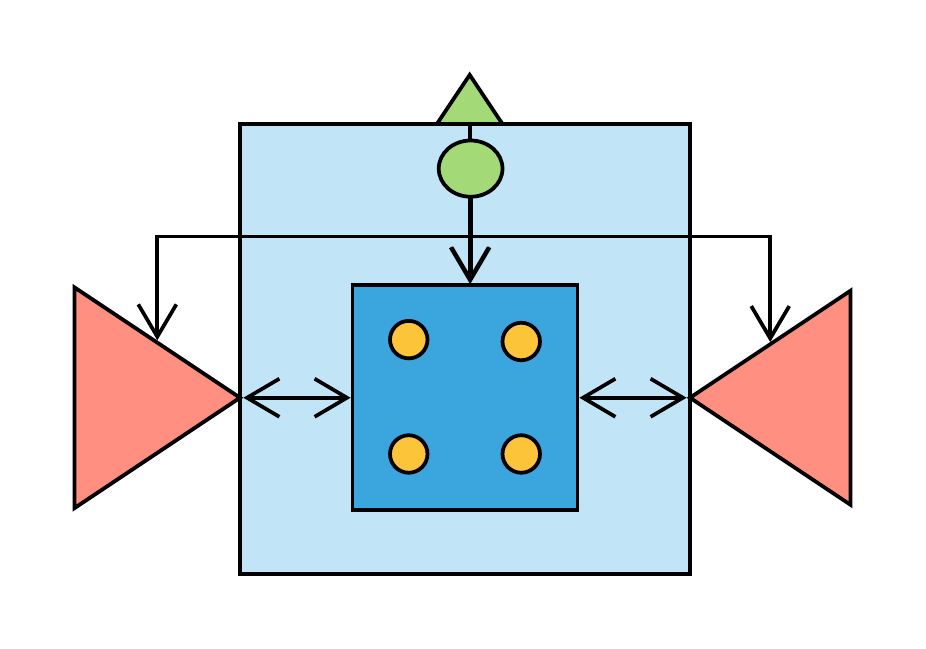}
\caption{Animat architecture. The green triangle/circle marks the sensor for the wall/other animats, the yellow circles mark hidden nodes and red triangles mark the motors. Sensors only connect to the hidden nodes and motors in a feedforward manner. Hidden elements and motors can feed back to all other nodes except sensors.}
\label{fig:animat}
\end{figure}

Each animat used in this simulation contains a set of 2 sensors (one for walls, one to detect other animats), 2 motors, and 4 hidden memory nodes. Figure \ref{fig:animat} shows a schematic of the animats' architecture.

Each animat has two kinds of sensors: one sensor detecting obstacles, here the walls, (green triangle) and one sensor detecting other animats (green circle). Both types of sensors have a range of 1 unit directly in front of the animat. Animats are built with feedback motors. The motor elements can thus also act as memory just as the hidden nodes, which means that the current motor state at $t$ can be causal for the future `brain' state $t+1$ (this is easily observable in the example wiring diagram below, Figure \ref{fig:wire}). All nodes in the network can have two states, 1 and 0. A sensor switches to 1 if it detects an obstacle or animat, respectively. The movement model contains four possible states mapped by a 2-bit tuple $M = (m_l,m_r)$ where $m_l$ and $m_r$ model the left and right motor. $M = (1, 0)$ implies that only the left motor is active and therefore the animat turns left
. The same holds for $M = (0, 1)$ in which the animat turns right. $M = (0, 0)$ indicates a static animat and $M = (1, 1)$ means that it moves forward. 

Designing the animats with limited sensors and motors increases relative task difficulty, which has to be compensated by more complex internal states \cite{Albantakis2014} and, for the same reason, should also facilitate the evolution of cooperation. Using the sensor data as the environment's representation, an animat has an internal representation stored in its artificial brain. There are several common types of such models: the brain could simply be a manually coded function, an ANN \cite{Stanley2005a}, a simple finite state machine \cite{Konig2009}, or a MB \cite{edmund2011}. In our case, the focus is on elucidating the animat's behavior while observing its internal and external states. Therefore, a simple and representable cognitive system was required. This is why we chose to implement MBs. Moreover, MBs emulate principles of neocortical function due to their strong embodiment of neuronal cognitive processes \cite{Marstaller2013}. Future work should also consider other methods like ANN and Finite State Machines.

\begin{figure}[!htp]
\includegraphics[width=0.3\textwidth]{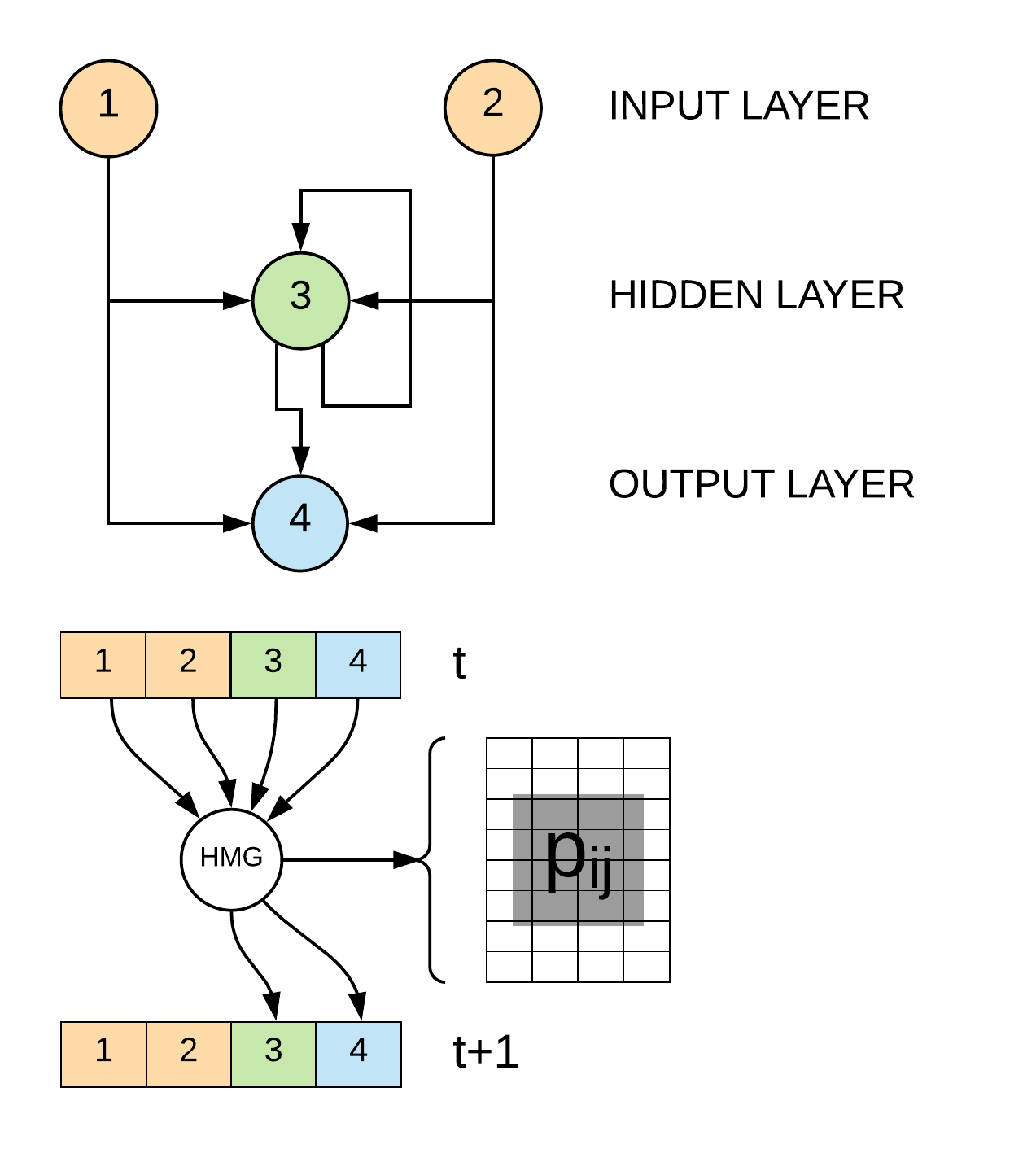}
\caption{A MB \cite{edmund2011} is composed of nodes, HMGs and their connections. The HMGs specify the mechanisms to transform a brain state at time $t$ to the future state at $t+1$, e.g. by fixed probabilities (indicated by $p_{ij}$). The effective brain connectivity between nodes (upper diagram) is derived from the nodes' hidden connections to and from the HMG.}
\label{fig:mb}
\end{figure}

A MB is composed of a set of nodes with a finite set of states, which have temporal dependencies. The nodes' state-dependent update rules are implemented with the support of \textit{Hidden Markov Gates (HMG)}, which indirectly connect the different nodes in the MB. Figure \ref{fig:mb} shows an example architecture of a 4-node MB with one HMG. Each node can have a state (e.g. $1$ or $0$). A set of input nodes is connected to a HMG. Inside the HMG there could be a lookup table, or any other mechanism, transforming the inputs at $t$ into an output at $t+1$. The HMG's output is written to a set of output nodes, which could determine a motor state and/or be used as memory in the next time step. In this study, we exclusively used deterministic lookup tables to specify the HMGs' input-output functions. The HMG's input-output functions and their inputs and outputs are encoded via a genome consisting of a string of integer values. At every generation, each locus in the genome has a probability of mutation; small sections of the genome also have a probability of being deleted or duplicated \cite{Hintze2017, edmund2011} (all parameters as in \cite{CliffordBohmNitashCG2017}, see also supplementary material A.1).

Note that there is no active communication between the animats in this study. This means that the agents do not share information between each other, as, for instance, two organisms having a dialog. Animats can only sense whether another animat is directly in front of their position (or not). 

\subsection{Grid-Based Environment and the Challenge to Move Through the Gate}
\label{sec:fitness}

In this work, we were interested in swarms and the evolution of swarm behavior, not just single animats as in \cite{edmund2011, Marstaller2013, Albantakis2014}. Accordingly, multiple animats were placed in the environment simultaneously. Here, a swarm contained only clones, meaning each animat had the same genome and thus MB. Swarm size stayed constant across generations during each evolution.  5 different swarm sizes were tested.  This made it possible to investigate dependencies between swarm size and the animats' (swarm) behavior. We distributed the swarm sizes uniformly up to a maximum of 72 animats, corresponding to all predefined starting positions in the task environment (see below): 

\begin{enumerate}
\item $G_{1.00}$: 72 animats, the total number of available starting positions (constrained by the environment design).
\item $G_{0.75}$: 54 animats, 75 percent of the starting positions.
\item $G_{0.50}$: 36 animats, 50 percent of the starting positions.
\item $G_{0.25}$: 18 animats, 25 percent of the starting positions.
\item $G_{single}$: Only one\footnote{Obviously, one animat cannot form a swarm. We included this condition here for comparison and treated it equivalently to simplify formulations.} animat is placed in the environment.

\end{enumerate} 

\begin{figure}[!htp]
\includegraphics[width=0.25\textwidth]{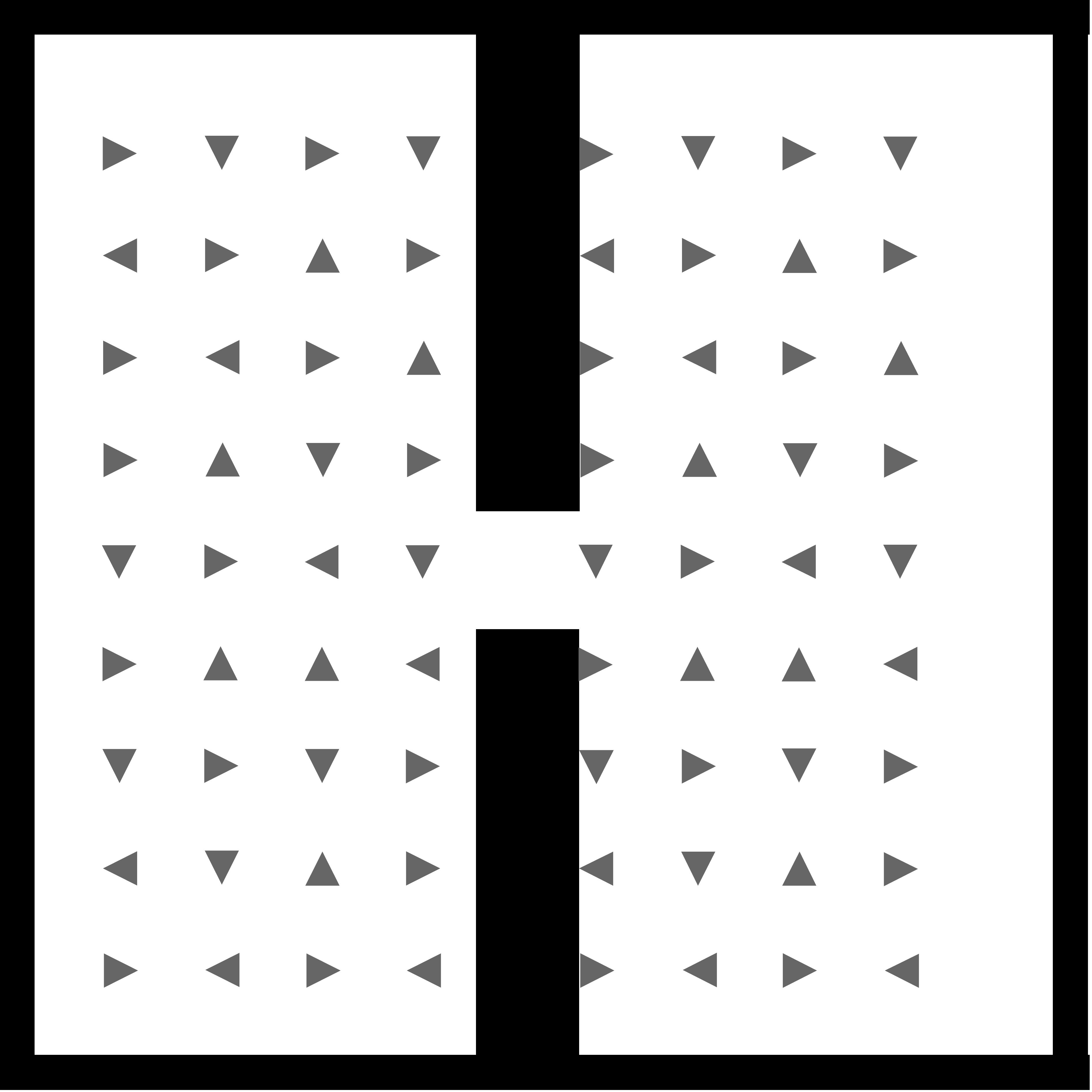}
\caption{Grid-based environment design. The environment contains two rooms, which are connected by a narrow gate. There are 72 fixed starting positions for the animats, onto which the animats are placed at random, facing in random directions (up, down, left, right).}
\label{fig:world}
\end{figure}

The grid-based, 2-dimensional environment is designed to have $32 \times 32$ units (Figure \ref{fig:world}). At initiation, the animats were placed randomly without overlap on 72 predefined starting positions as marked in Figure \ref{fig:world} by gray triangles. The environment was partitioned into two rooms connected only by a narrow gate, which the animats were supposed to cross frequently as part of their task. The design was inspired by the work of Koenig et al. \cite{Konig2009}.
 
The task environment is designed to pose a multi-objective problem. On the one hand, an animat receives reward if it is able to travel through the gate. On the other hand, it gets a penalty if it collides with other animats (occupying the same location). 
In this paper, we select a tailored weighted sum approach to solve the multi-objective problem. In order to select the weights, we consider the following feature.
Since collisions between two animats are much more likely than crossing the gate, the collision penalty is set to a lower value than the reward for traveling through the gate, while still allowing for optimal behavior (0.075 compared to 1). If the penalty is chosen too low, all animats would learn to center around the gate area, colliding with each other all the time. If the penalty is too high, all animats would adapt to not move at all (data not shown). Additionally, since it is not desirable for animats to crowd around the gate area, a timeout of $100$ time steps is implemented until an animat can receive further rewards after crossing the gate once. Each trial had a total duration $T$ of $500$ time steps\footnote{With less time steps, early generations would hardly encounter the gate, impeding evolution.}. This time-out period could be interpreted as simulating a requirement to make the way back to the organism's nest and also promotes the evolution of unified behavior.

\subsection{Setup of the Genetic Algorithm}
\label{sec:ga}

In the following we define the mathematical notation and equations for the fitness function. Table \ref{tab:fitsym} lists all parameters and variables. In Equation (1) the fitness for a single animat in the environment is defined. Equation (2) specifies the overall fitness of the genome, or MB, which is also used for the selection process in the \textit{genetic algorithm (GA)} of MABE. At each generation, we tested the swarm's genome $30$ times in the environment ($|R| = 30$), with different random starting parameters (starting position and orientation) to obtain robust fitness values for each genome. In each of these $30$ trials, we randomly picked a single animat out of the swarm\footnote{This was done in order to maintain the same sample size across conditions with different swarm sizes. In this way, we avoided any bias in the fitness evolution merely due to differences in the variability of the fitness values across generations. Averaging across the entire swarm in condition $G_{1.00}$, for example, would eliminate any variability in fitness due to the random starting positions.} and averaged across their fitness values to obtain the overall fitness assigned to the genome.
\begin{table}[!htb] 
\caption{Definition of the Mathematical Notation for the Fitness Function}
\begin{tabular}{ll}
	$a$					& Identifier of a single animat $a$, where $a \in \mathbb{N}$  \\
	$A$					& The set of all animats $a$ in a trial, i.e. a swarm\\
	$R$					& The set of all trials $R_i$ an animat is tested in  \\
	$f(a)$				& The fitness of a single animat $a$ \\
	$F(A,R)$			& The average fitness of a genome \\
	                 & across all trials $R$ \\
	$\mathrm{randA}(A,R_i)$		& Picks a random animat $a$ from the swarm $A$ \\
	                 & depending on the trial $R_i$ \\
	$g(a,t_a,t_b)$	& Returns the count of gate-crossings between time   \\
  	& $t_a$ and time $t_b$ for a single animat $a$ \\
	$c(x,y,t)$		& Returns the count of animats at a \\
	                & specific position $(x, y)$ at time $t$ \\
	$t$					& A single time step $t$, where $t \in T$ and $t \in \mathbb{N}$ \\
	$T$					& Trial duration, i.e. the number of all time steps $t$ \\
	   				& in a trial\\
	$x(a), y(a)$		& Returns the $x$ and $y$ position of animat $a$
\end{tabular}
\label{tab:fitsym}
\end{table}

\begin{dmath}
f(a) =  \sum_{t=0}^{T-1} \begin{cases}
	1 & g(a,t, t+1) = 1 \text{ and } g(a, t-100,t) = 0 \\
	0 & \textit{otherwise}

\end{cases}  - \sum_{t=0}^{T}\begin{cases}
	0.075 	& c(x(a),y(a),t) > 1 \\
	0 		& \textit{otherwise}
\end{cases}
\end{dmath}

\begin{dmath}
F(A,R) =  \dfrac{\sum_{i=1}^{|R|} f(\text{randA}(A,R_i))}{|R|}
\end{dmath}

A single evolution experiment was run for 10,000 generations. At each generation, a population of 100 genomes was evaluated, encoding the animats' MBs. After each generation, a set of 100 genomes was selected (with the possibility for duplicates) to enter the next generation based on their fitness values and the selection rules of the GA. These genomes were then mutated according to the probabilities specified in A.1 \cite{CliffordBohmNitashCG2017}. Note that the genome population should not be confused with the swarm size: a swarm is a set of clones with identical genomes and thus MBs. The population of genomes corresponds to the pool for selection. For each of the 5 $G_i$ conditions we ran $30$ evolution experiments with different random seeds.  

Statistical differences were evaluated using a Kruskal-Wallis test, the non-parametric equivalent of a one-way ANOVA. Differences between pairs of task conditions reported in the results section were assessed by post-hoc Mann-Whitney U tests. See supplementary material A.2 for detailed comparisons.


\section{Results}
To address our research questions, we performed a multi-level analysis to evaluate the evolved genomes resulting from the GA. First, we compared the average fitness evolution of the 30 evolution experiments per condition of the different swarm sizes $G_i$. Second, we investigated the movement patterns of all agents in a swarm to answer whether the animats evolved swarm behavior. Third, we evaluated the animats' generalizability across swarm sizes. An animat is generalizable if it performs at high fitness when tested with different swarm sizes as it was trained in. We also report qualitative observations about behavioral differences between the various swarm sizes $G_i$. Finally, we applied a simple graph-theoretical measure to the animats' MBs as a proxy for brain complexity.

\subsection{Evolution of Fitness}
\begin{figure}[!htp]
\includegraphics[width=0.45\textwidth]{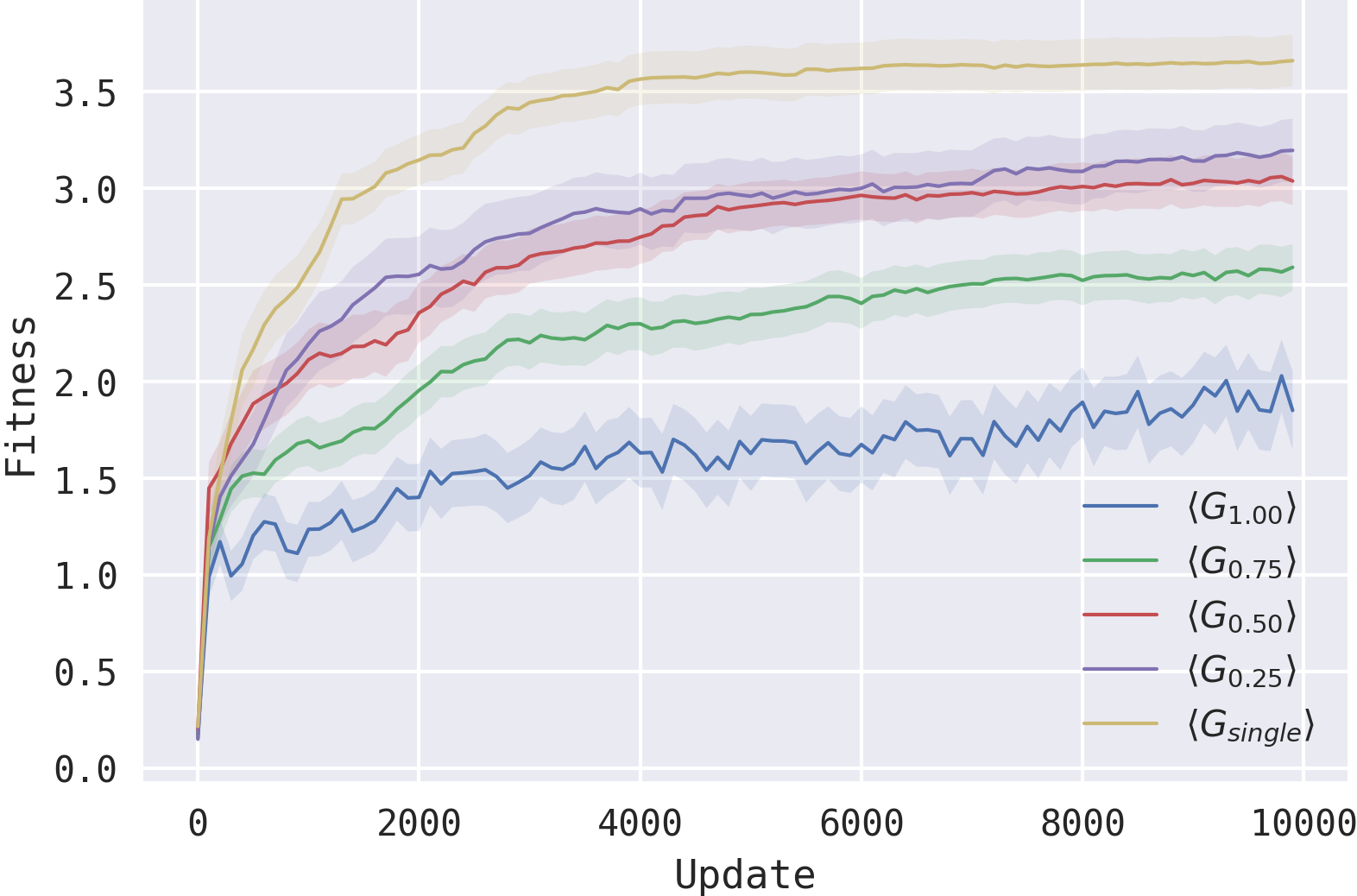}
\caption{Task fitness averaged across 30 evolutions of the five different configurations $G_i$. The overall fitness increases during evolution, while the final fitness decreases with swarm size. The shaded area indicates the Standard Error of the Mean (SEM).}
\label{fig:lod}
\end{figure}

First, we analyzed the evolution of fitness across all test groups $G_i$. As it can be observed in Figure \ref{fig:lod}, task fitness is strongly dependent on swarm size: the smaller the swarm the steeper the curve and the higher the final evolved task fitness. All groups differed significantly ($p<0.05$) from each other in their final fitness (see A.2 for details). This result shows that the task, in general, can be solved without any cooperation and actually becomes more difficult for larger swarm sizes since colliding with other animats results in penalty (ref. section \ref{sec:fitness}).

\subsection{Observation of Swarm Behavior}

\begin{figure}[!htp]
\includegraphics[width=0.45\textwidth]{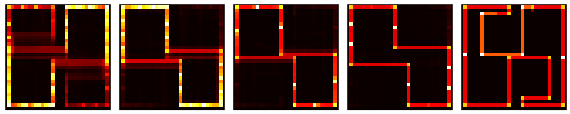}
\caption{Heatmaps of the best genome of all $G_i^{10k}$. From left: $G_{1.00}^{10k}$, $G_{0.75}^{10k}$, $G_{0.50}^{10k}$, $G_{0.25}^{10k}$, $G_{single}^{10k}$. Color indicates occupation density during the duration of one trial. Black areas were never visited, while red to white areas mark low to high density. Yellow/white cells and areas indicate spots where the animats turned or stalled frequently. This `stop-and-go' behavior was more common for animats evolved in large swarms.}
\label{fig:hm}
\end{figure}

Secondly, we tested if the animats developed swarm behavior or only independent movements. For this purpose, we generated heat-maps highlighting the movement patterns of the animats during their trial.  Figure \ref{fig:hm} shows the 5 heat-maps of the best genomes for each swarm size condition at the final generation in the GA ($G^{10k}_i$). We also generated and inspected animations of the swarm's evolved behavior (final generation) for each of the 30 evolutions per condition. The most common movement patterns fit a `stop-and-go' wall-following strategy, as in the study by Koenig et al. \cite{Konig2009}. According to Pacala et al. \cite{Pacala1996}, such a strategy qualifies as swarm behavior as it is a result of social interactions and interaction with the environment. This would mean that also the interaction with the wall is part of the swarm-behavior. For example, organisms in $G^{10k}_{single}$ tended to receive less stimuli from the wall as shown below in Figure \ref{fig:states_all}.

Most swarms with good fitness evolved such a wall-following strategy, but diversity in the movement patterns was also observed, particularly for animats with a lower final fitness. Nevertheless, only animats in $G_{single}$ evolved high fitness using qualitatively different strategies. By distinguishing between (dark) red and yellow/white cells it is possible to observe whether the swarm is moving steadily or not. As the examples in Figure \ref{fig:hm} show, big swarms moved slower along the walls, while smaller swarms only exhibited a few halting or turning points, particularly in the corners. 

\subsection{Generalizability of Animats}
\label{sec:robust}
\begin{figure}[!htp]
\includegraphics[width=0.45\textwidth]{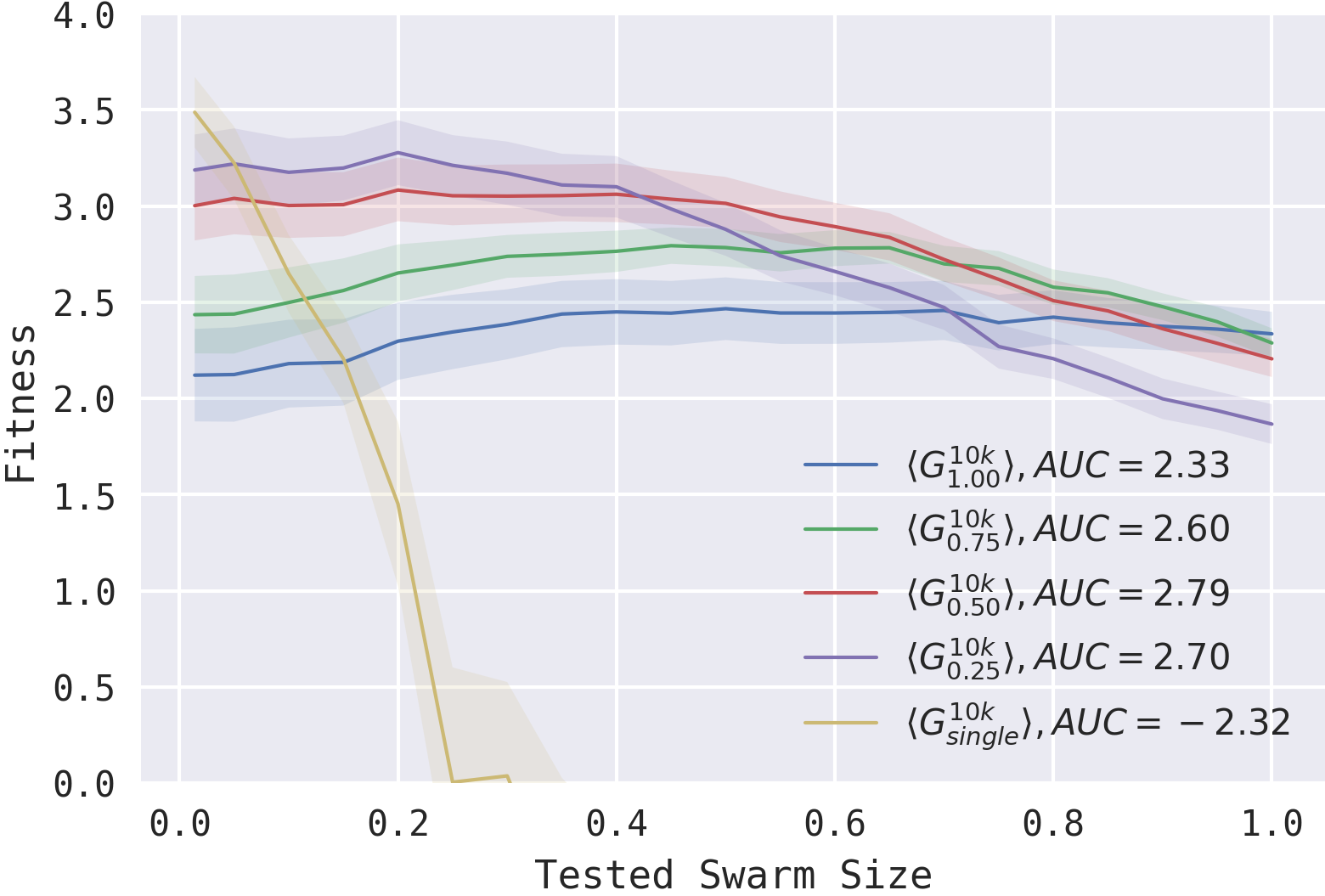}
\caption{Average performance of the final generation of each evolution experiment grouped by swarm size during evolution $G^{10k}_i$ when tested at different trial swarm sizes $ [100\%, 95\%, ... , 10\%, 5\%, \text{single} ]$.}
\label{fig:group_robustness}
\end{figure}

To test the generalizability of the evolved animats, we tested the final generation of animats of all conditions ($G^{10k}_i$) in swarm sizes other than the one they evolved to, specifically, at \\ $[100\%, 95\%, ... , 10\%, 5\%, \text{single} ]$ of the maximal swarm size of 72 individuals. We then compared the robustness of their performance across swarm sizes to observe their generalizability (Figure \ref{fig:group_robustness}).  

While $G^{10k}_{single}$ animats failed to maintain their fitness within a swarm, all animats that were evolved in an actual swarm demonstrated a fair amount of generalizability. We quantified the fitness robustness of the different animat conditions $G^{10k}_i$ by calculating the \textit{Area under the curve (AUC)}, which is largest for the animats in $G^{10k}_{0.50}$, followed by $G^{10k}_{0.25}$. 
 All conditions except for $G^{10k}_{0.50}$ and $G^{10k}_{0.25}$ ($p=0.1548$) differ significantly ($p<0.05$) from each other (see A.2 for details).
Note also, that $G^{10k}_{0.50}$ showed comparable fitness values to all other conditions except $G^{10k}_{single}$ in their original evolutionary swarm size.
This suggests that adapting to intermediate swarm sizes may provide an advantage under changing environmental conditions, such as variation in swarm size due to rare environmental events. Our findings are also in line with Brown \cite{Brown1982} that too low and too high swarm density is negative for the overall swarm performance and respectively for the individual organism as well. 

\subsection{The Cognitive Processes of the Animats}
\label{sec:states}

\begin{figure}[!htp]
\includegraphics[width=0.45\textwidth]{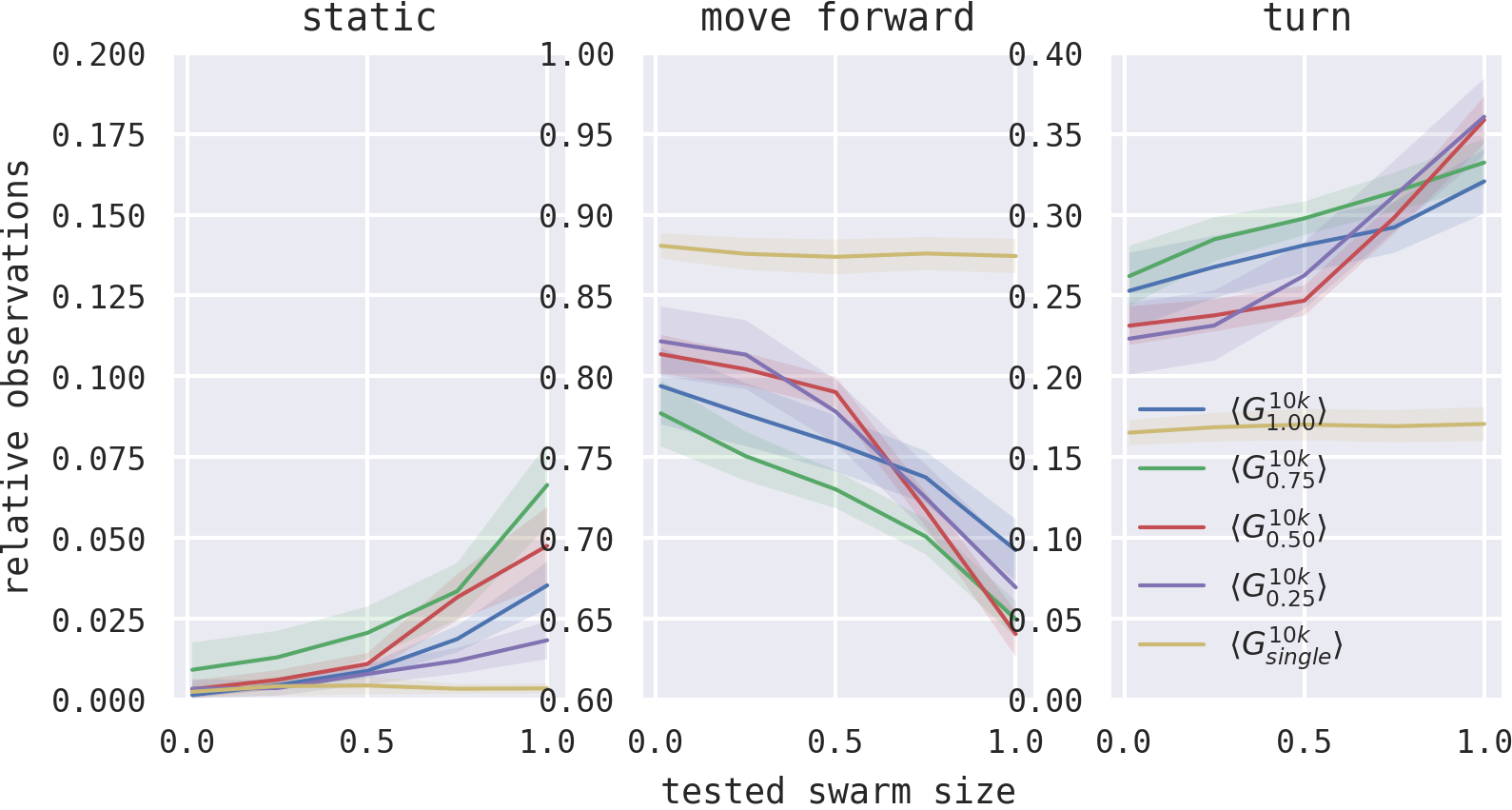}
\caption{Number of movements, turns, and no-movements scaled by trial duration ($T = 500$ time steps) for all $G^{10k}_i$ plotted against the different test population sizes.}
%
%
\label{fig:states_flat}
\end{figure}

To identify regularities regarding the decision-rules in the respective swarm size conditions $G_i$ we evaluated the animats' cognitive processes while performing the task. First, we evaluated the frequency of the animats' various motor states (Figure \ref{fig:states_flat}) of all final animats $ G^{10k}_i $ while performing the task in different swarm sizes (same data as in Figure \ref{fig:group_robustness}). Here, variation across conditions indicates cognitive flexibility. What is more, these data also allowed us to differentiate whether being part of the swarm made the animats act in a certain way, or if, in turn, the swarm is merely the result of individual reactions. For $G^{10k}_{single}$ it should be obvious that individual animats were not influenced by the swarm and if seeming swarm behavior was observable it was not due to interactions with other animats. This is also supported by the data: $G^{10k}_{single}$ shows no variation in its motor responses across conditions. By contrast, animats evolved in swarms adapted their behavior dependent on the swarm size they were placed in. In particular conditions $G_{0.50}^{10k}$, which demonstrated the greatest generalizability (Figure \ref{fig:group_robustness}), also had the most dynamic reaction to the different swarm sizes. 

\begin{figure}[!htp]
\includegraphics[width=0.35\textwidth]{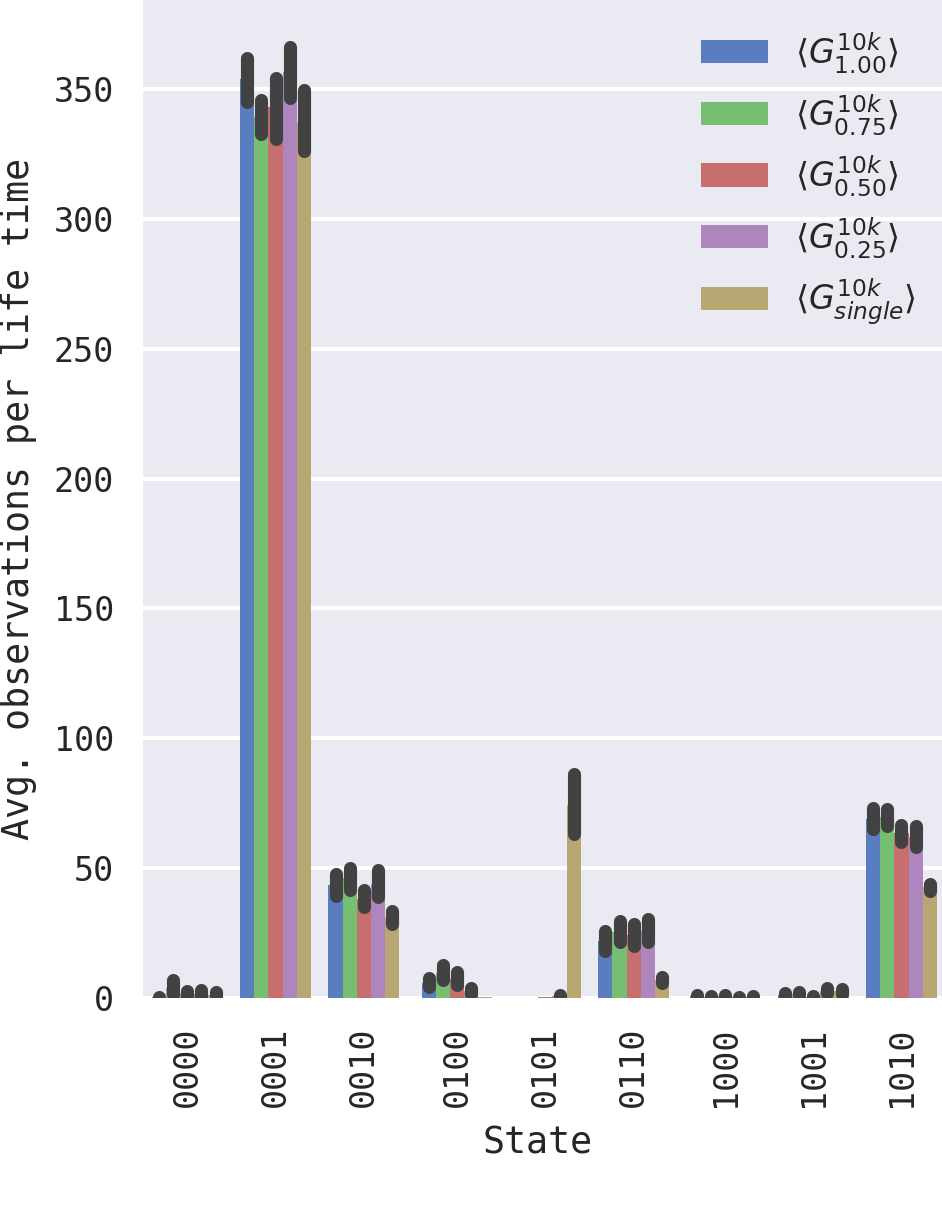}
\caption{The average number of times an animat enters a specific state transition, grouped by $G_i^{10k}$. The tuple is coded as \textit{(wall sensed, animat sensed, turn, move forward)}. Black bars mark the bootstrapped confidence intervals of the number of observations across evolutions.}
\label{fig:states_all}
\end{figure}

Second, we measured how often specific sensory-motor state transitions could be observed (Figure \ref{fig:states_all}). Specifically, we recorded which actions at $t+1$ followed a particular input at $t_{x}$. This corresponds to a complete external representation of the animats, i.e. their input-output behavior. Having two sensors and two motors, there are $2^4$ possible external states. These were condensed to 4 bits in order to capture the following information: (1) The animat senses a wall, (2) the animat senses another animat, (3) the animat turns left or right, and (4) the animat moves forward (2 motors active). Note that, because of the nature of the task environment, instances such as sensing a wall and an animat at the same time or turning and moving forward at the same time are impossible and thus not considered, leaving $9$ different state transitions to be evaluated. As an example,  $0101$ indicates that an animat sees another animat at $t$ and moves forward at $t_{x+i}$.
%

Figure \ref{fig:states_all} shows the cumulative state transition count during a trial, averaged across the 30 evolutions for each $G_i^{10k}$. Since the first-order statistics of sensor and motor states depend on the size of the swarm during a trial, each animat evolved in a particular $G_i$ was tested on each swarm size in the interval of $[100\%, 75\%, 50\%, 25\%, single]$ of the environment's maximal capacity (72 animats). This means that we counted all occurred transitions in $5$ different trials of different swarm size. Moving forward without previously spotting another animat is by far the most frequent action in all conditions. It can also be observed that animats in $G^{10k}_{single}$ simply ignore other animats colliding with them despite the penalty ($0101$), while the others instead evolved to turn ($0110$) in such a situation. Finally, animats in $G^{10k}_{single}$ seem to rely less on sensing the wall for guidance ($1010$), which could mean that they overall react less to their environment but rather use memory to solve the task (see also the heatmap example in Figure \ref{fig:hm}).
 
\begin{figure}[!htp]
\includegraphics[width=0.48\textwidth]{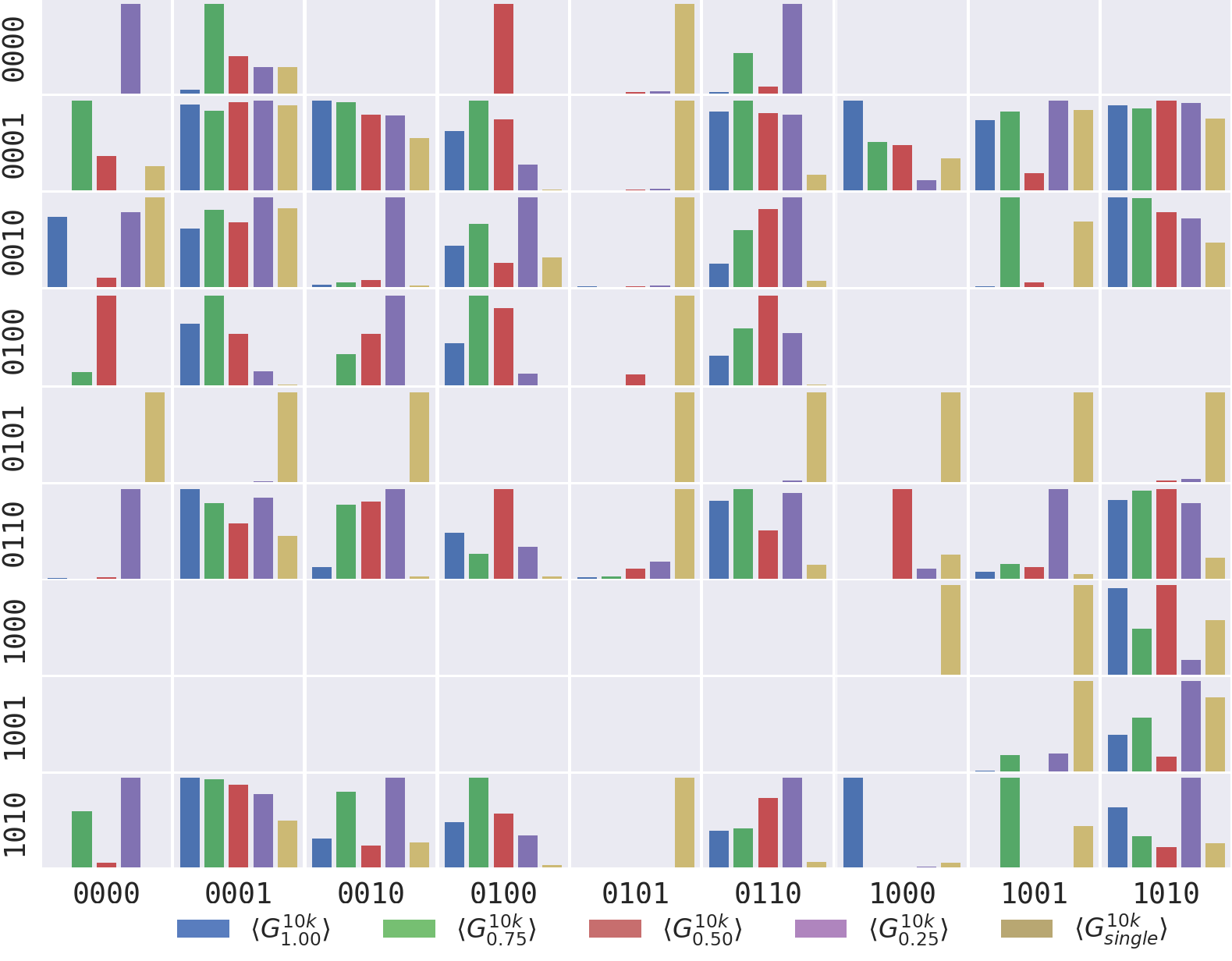}
\caption{External states of all $G^{10k}_i$ over three time steps. The axes are labeled with the state (see text for details). One tile shows how often the state transitions occur per $G^{10k}_i$. The values of each single tile are scaled between $1$ and $0$, where $1$ is the maximum probability to enter that transition and $0$ simply the zero probability across conditions $G^{10k}_i$.}   
\label{fig:tpm}
\end{figure}

To observe more detailed differences between the respective conditions than in Figure \ref{fig:states_all}, we took one time step more into account and considered inputs at $t$, the reactions at $t+1$, the new inputs at $t+1$ and the corresponding reactions at $t+2$. To visualize the data we generated a \textit{transition probability matrix (TPM)}  (Figure \ref{fig:tpm}). For the sake of readability, we limited the labels in the plot and will thus describe them here. On the x-axis there are all inputs/reactions at $t$/($t+1$), on the y-axis there are all inputs/reactions at ($t+1$)/$(t+2)$. In the $9 \times 9$ matrix one tile visualizes the scaled probabilities per 3-time-step transition and $G^{10k}_i$. Since we were more interested in differences across conditions than the absolute number of transitions, we scaled the bars according to their maximum and minimum values over all swarm conditions $G^{10k}_i$, to better spot possible differences. Furthermore, values were averaged over all $30$ different evolutions per condition, tested, as above, in 5 trials with different swarm sizes.  

Animats in $G_{0.50}^{10k}$ stand out regarding their generalizability. Reviewing the TPM one can observe that such animats stayed static more often as the average, especially when spotting an animat or a wall, and also remained static in the following time step. $G_{0.75}^{10k}$, who would be static more often, rather stays static when sensing nothing at all. This also supports our conclusion from Figure \ref{fig:states_flat} that the behavior of animats in $G_{0.50}^{10k}$ was influenced most by sensing other animats and the environment. While it is observable that animats in $G_{single}^{10k}$ always tried to move forward (spotting an animat or not), which had no negative effect in their original evolution environment, we observed that animats in $G_{1.00}^{10k}$ turned more often, even if they have no specific input.

\subsection{Brain Complexity}

\begin{figure}[!htp]
\includegraphics[width=0.45\textwidth]{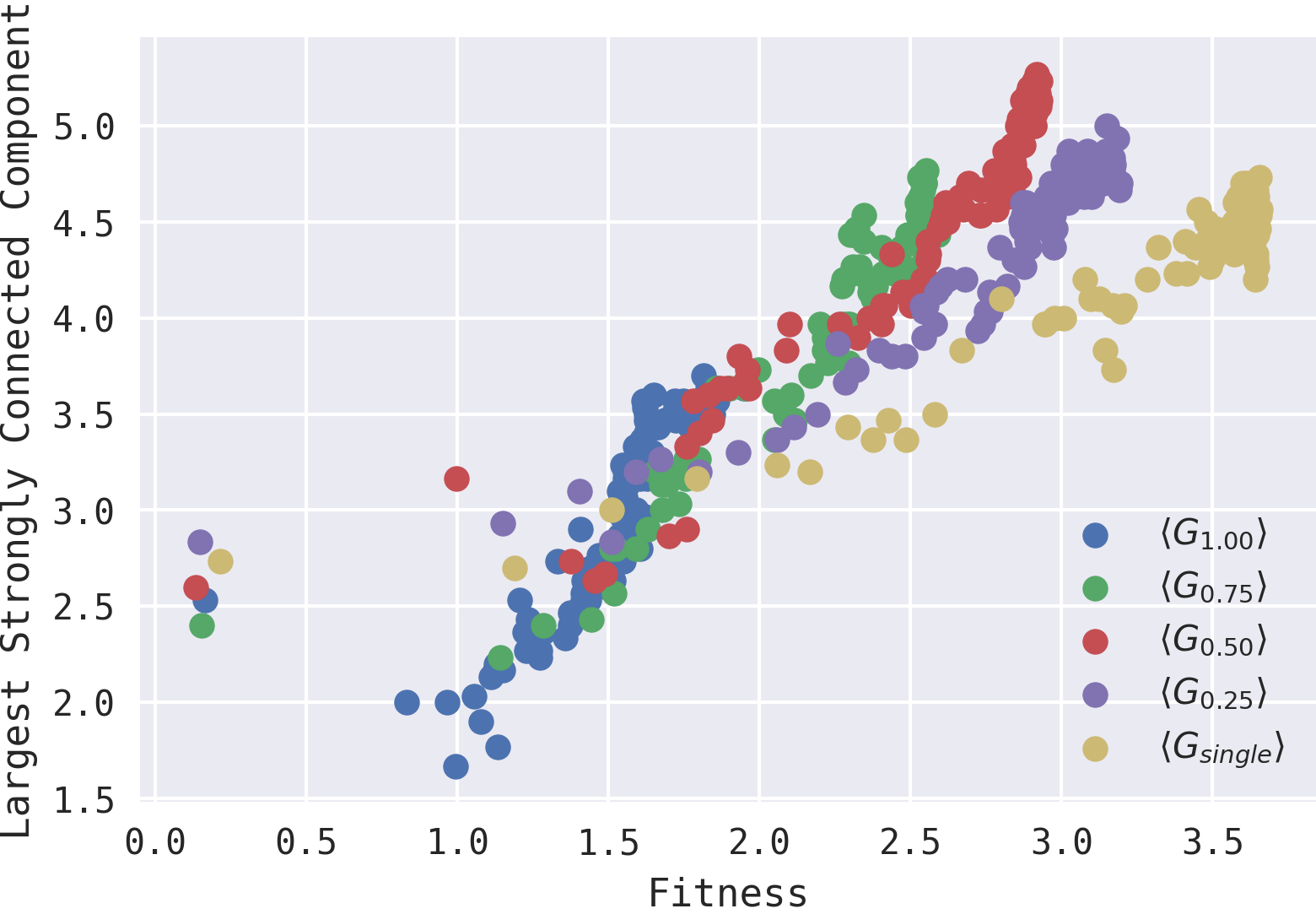}
\caption{Fitness plotted against the LSCC of the animat's brain, which we used as a proxy for brain complexity. One dot is the average LSCC over 30 experiments per generation evaluated at every $100^{th}$ generation)}.
\label{fig:lscc}
\end{figure}

The results presented above indicate that $G_{0.50}$ animats evolved the most generalizable behavior. Apart from the animats' externally observable input-output behavior, we also wanted to take their internal structure into account. The node connectivity in a MB can be modeled as a directed graph. As a simple graph theoretical measure of brain complexity, we thus used the \textit{Largest Strongly Connected Component (LSCC)}, which is also a simple measure of a graph's integration\footnote{Other, less significant, graph theoretic measures, which show the same trend, can be found in the supplementary material A.3.}. Additionally, we would take the LSCC or similar metrics into account to determine, which parts of the MB influences the future brain states most. As shown in Figure \ref{fig:lscc}, animats acting alone or in large groups tend to evolve significantly ($p<0.05)$ less complex brains even at high levels of fitness (see supplementary material A.3). Our assumption is that for $G_{single}$ the environment was comparatively simple and rules to achieve high fitness were easier to find. Animats evolved with $G_{1.00}$ could rely on sensing other animats with a high probability, which could serve as an orientation. $G_{0.50}$ evolved the most complex brain structures, which relates to our previous observations of the comparatively high behavioral complexity and generalizability of this group. 


\begin{figure}[!htp]
\includegraphics[width=0.25\textwidth]{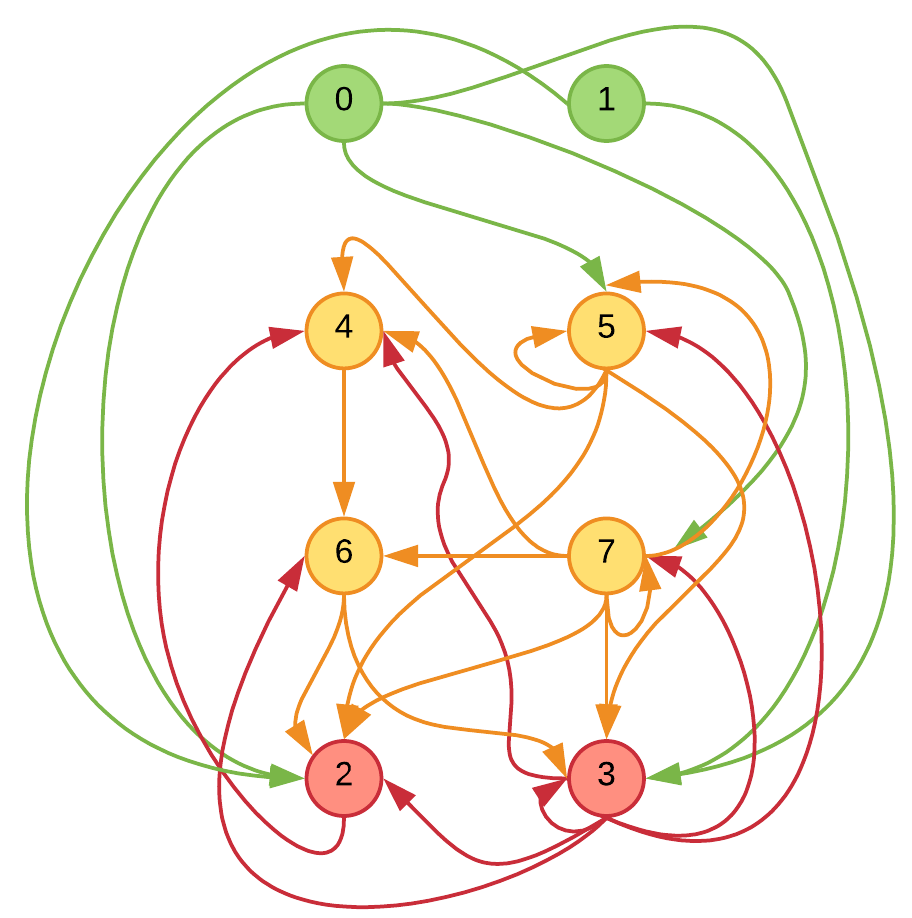}
\caption{Wiring Diagram of the best final genome in $G^{10k}_{0.50}$. The green circles mark sensor nodes, the yellow circles mark hidden nodes and the red circles mark motor nodes.}
\label{fig:wire}
\end{figure}

Figure \ref{fig:wire} shows the wiring diagram of the best animat in $G_{0.50}^{10k}$ with an average task fitness of $F(A) = 3.8$. The animat has feed-forward sensors. All other nodes can feed back to each other, except to the sensors. This animat thus has the largest possible LSCC of 6. Feedback-loops are also an indicator for memory in a MB.

%
%
%

\section{Future Work}

The present results were obtained from one particular task environment. To build empirical strength, implementing more difficult and diverse tasks will be required, e.g., the predator-prey scenario used in earlier works by Olson et al. and Miikkulainen et al. \cite{Olson2016, Miikkulainen2012}. Additionally, it is important to investigate variations in the animats' design to determine how the kind and number of sensors influence their behavior. In this work, animats were evolved and tested in an isolated manner. This means that a swarm only contained clones of one animat genome, which was necessary to make a first, specific evaluation. For future work, the effect of diverse swarms should be considered, which might increase the computational performance and make the simulated experiment more realistic. Finally, the development of more rigorous statistical analyses of the animats' external and internal state transitions is current work in progress.

\section{Conclusion}

Evaluating the detailed behavior and interactions of organisms in a simulated swarm is an open field of research. In this work, we addressed the effect of swarm size during evolution in a 2-dimensional spatial navigation task in a framework in which animats with Markov Brains were trained using a Genetic Algorithm. We, moreover, evaluated to what extent the resulting animats would be generalizable (testing animats in swarm sizes different from the one they evolved in). Furthermore, we focused on the evaluation of the animats' swarm behavior and its flexibility when faced with swarms of different sizes. We found that swarm size matters in the evolution of swarm behavior. Even if the task did not require cooperation, animats reacted to other animats non-egoistically in their decisions and formed swarm behavior. Our observation is that animats evolved in very large or very small swarms were less generalizable to other swarm sizes and showed less flexibility in their behavior. We assume that individuals in large swarms primarily acted to avoid collisions and the associated penalty, while animats in small swarms had less incentive to develop proper reactions to encountering other animats. Overall, our results suggest that animats evolved at intermediate swarm sizes may have adaptive advantages due to their more generalizable and flexible behavior, which is also reflected in their higher relative brain complexity.

\begin{acks}
We would like to thank Arend Hintze and Clifford Bohm at Michigan State University for their early advice on this project, sharing their extensive experience with the evolution of artificial organisms, and, in particular, their help with the MABE framework. 
\end{acks}

\bibliographystyle{ACM-Reference-Format}
\bibliography{bibliography}

\newpage

\appendix

\section{Supplementary Material}

\subsection{MABE}
\label{sec:a_mabe}
The following settings were used to configure the GA: 
\begin{enumerate}
    \item Settings for the Genome
\begin{enumerate}
    \item Type: Circular
    \item Alphabet Size: 256
    \item Sites Type: char
    \item Initial Size: 5,000
    \item Mutation Point Rate: 0.005
    \item Mutation Copy/Delete Rate: 0.00002
    \item Minimal Mutation Copy/Delete Size: 128
    \item Maximum Mutation Copy/Delete Size: 512
    \item Minimal Size: 2,000
    \item Maximal Size: 20,000 
\end{enumerate}

\item Settings for the Markov Brain:
\begin{enumerate}
    \item Type of Gates: Deterministic
    \item Range of Inputs/Outputs per Gate: 1-4
\end{enumerate}

\item Settings for the Optimizer:
\begin{enumerate}
    \item Type of Optimizer: Tournament
    \item Tournament Size: 5
    \item Population Size: 100
    \item Elitism: No
\end{enumerate}

\end{enumerate}
\subsection{Significance Tests}
\label{sec:a_sign}
Our conclusions are mainly formed using the evolution of fitness values, the fitness values of final evolved animats performing in different group sizes and the brain structures of final evolved animats. The data in the evolution of fitness contains fitness values of 30 random evolutions. For the significance test we only took the last generation into account. The data in the generalizabilty of the animats contains fitness values of final evolved animats, which were tested in groups of different size (see Figure \ref{fig:group_robustness}). The data to calculate the brain complexity results from the MB's connectivity matrix of all final evolved animats. In this section we present the significance tests of the analysis on this data. 

To test the significance, we first performed a Kruskal-Wallis test between all groups, followed by a Mann-Whitney-Test to test each pair of group. Since the Kruskal-Wallis was significant throughout all groups, we only provide the results of the Mann-Whitney-Test. 

\begin{table}[H]
\caption{Fitness Evolution Significance: p-values for the Mann-Whitney-Test testing the significance of the fitness of the last generation between the groups ($n_{ij} = 30$).}

\begin{tabular}{lrllll}
\toprule
 &  $G^{10k}_{1.00}$ & $G^{10k}_{0.75}$ & $G^{10k}_{0.50}$ & $G^{10k}_{0.25}$ \\
\midrule
    $G^{10k}_{0.75}$ & p-value &           0.0001 &                  &                  &                  \\
            &   $U$  & 192 & & & \\
    $G^{10k}_{0.50}$ &  p-value &          0.0000 &           0.0000 &                  &                  \\
            &   $U$  & 50 & 68 & & \\
   $G^{10k}_{0.25}$ & p-value &           0.0000 &           0.0000 &           0.0012 &                  \\
            &   $U$  & 101 & 118 & 225 & \\
$G^{10k}_{single}$ & p-value &           0.0000 &           0.0000 &           0.0000 &           0.0000 \\
            &   $U$  & 0 & 0 & 64 & 168 \\
\bottomrule
\end{tabular}
\end{table}


\begin{table}[H]
\caption{Significance of group size generalizability using the fitness of the best animats, which were tested in different group sizes (see Figure \ref{fig:group_robustness}):  p-values for the Mann-Whitney-Test testing the significance of the fitness generalizabilty between the groups ($n_{ij} = 6,300$).}
\begin{tabular}{lrllll}
\toprule{}  &  & $G^{10k}_{1.00}$ & $G^{10k}_{0.75}$ & $G^{10k}_{0.50}$ & $G^{10k}_{0.25}$ \\

\midrule
    $G^{10k}_{0.75}$ &   p-value & 0.0000 &    &    &   \\
            &   $U$  & 168,671 & & & \\
   $G^{10k}_{0.50}$ & p-value & 0.0000 &  0.0000 &     &      \\
            &   $U$  & 121,207 & 138,487 & & \\
    $G^{10k}_{0.25}$ & p-value &   0.0000 &           0.0000 &           \textbf{0.1548} &    \\
            &   $U$  & 149,688 & 170,764 & 191,888 & \\
  $G^{10k}_{0.single}$ & p-value &  0.0000 & 0.0000 & 0.0000 &  0.0000 \\
            &   $U$  & 84,366 & 74,146 & 66,428 & 69,566 \\
\bottomrule
\end{tabular}
\end{table}

\begin{table}[H]
\caption{Significance of the average LSCC values of the animat's MB in the last generation:  p-values for the Mann-Whitney-Test testing the significance of the average LSCC values between the groups ($n_{ij}=30$).}

\begin{tabular}{lrllll}
\toprule{}  &  & $G^{10k}_{1.00}$ & $G^{10k}_{0.75}$ & $G^{10k}_{0.50}$ & $G^{10k}_{0.25}$ \\

\midrule
    $G^{10k}_{0.75}$ &   p-value & \textbf{0.2464} &    &    &   \\
            &   $U$  & 405 & & & \\
   $G^{10k}_{0.50}$ & p-value & 0.0253 &  0.0042 &     &      \\
            &   $U$  & 328 & 282 & & \\
    $G^{10k}_{0.25}$ & p-value &   \textbf{0.3033} &\textbf{ 0.1099} & \textbf{0.0734} &    \\
            &   $U$  & 416 & 370 & 360 & \\
  $G^{10k}_{0.single}$ & p-value &  \textbf{0.0614} & \textbf{0.2262} & 0.0003 & 0.0197 \\
            &   $U$  & 350 & 401 & 226 & 316 \\
\bottomrule
\end{tabular}
\end{table}

\subsection{Alternative Graph Theory Measures}
\label{sec:a_gt}
We evaluated the brain complexity by using four different graph theoretic measures: LSCC, the average shortest path between any of the elements in the MB, the betweeness centrality (a measure of centrality based on the shortest path) and the average degree of the elements in the MB (the degree is the number of connected neighbors). This section displays the plots and significance values for the last three mentioned measures.

\begin{figure}[H]
\includegraphics[width=0.4\textwidth]{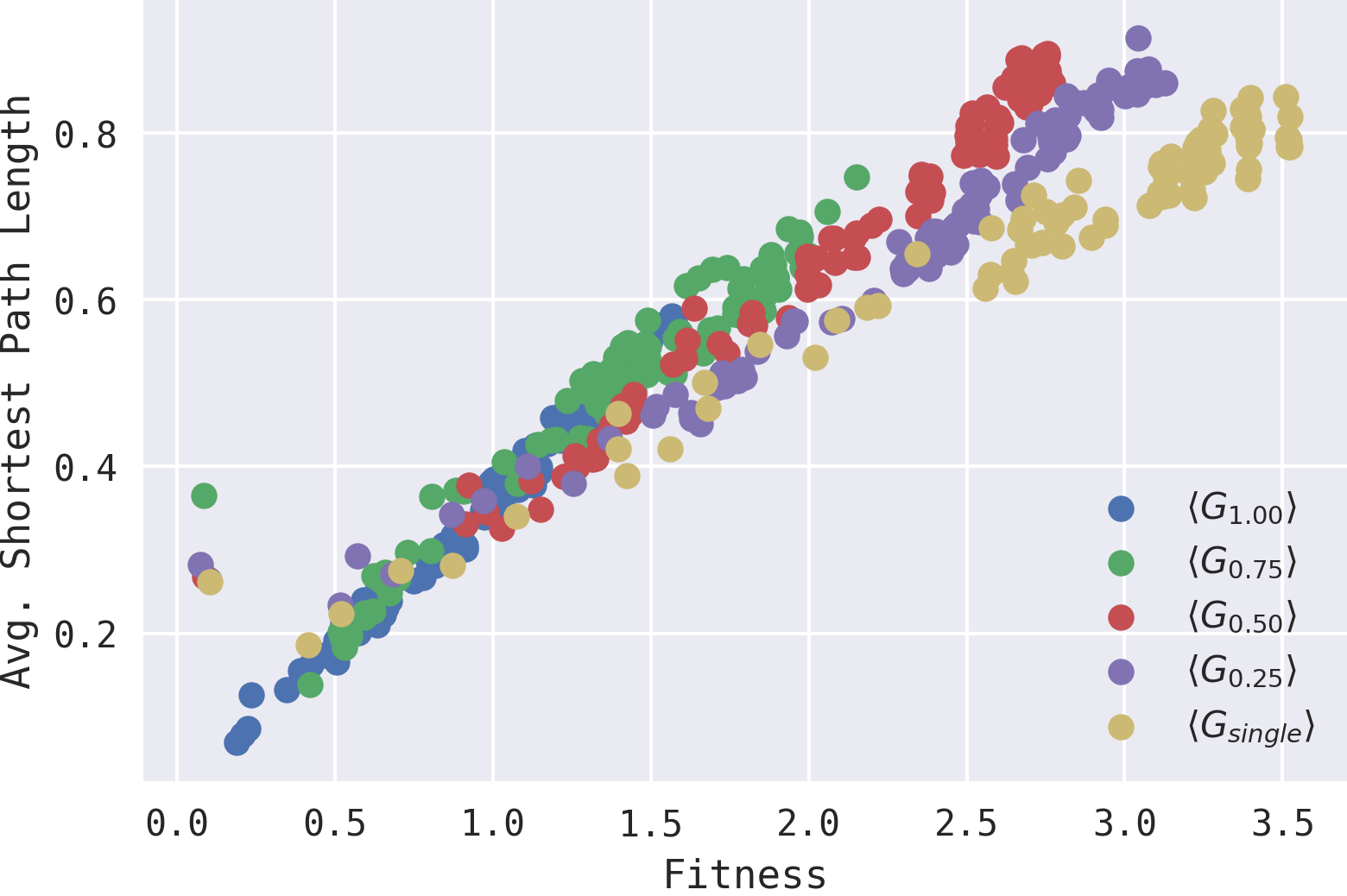}
\caption{Fitness plotted against the average shortest paths of the animat's brain mechanisms in the last generation. One data point is the average shortest path length over 30 experiments per generation evaluated at every $100^{th}$ generation).}
\label{fig:sPath}
\end{figure}

\begin{table}[H]
\caption{Significance of the average shortest path values of the animat's MB in the last generation:  p-values for the Mann-Whitney-Test testing the significance of the average shortest path values between the groups ($n_{ij}=30$).}

\begin{tabular}{lrllll}
\toprule{}  &  & $G^{10k}_{1.00}$ & $G^{10k}_{0.75}$ & $G^{10k}_{0.50}$ & $G^{10k}_{0.25}$ \\

\midrule
    $G^{10k}_{0.75}$ &   p-value & \textbf{0.2006 }&    &    &   \\
            &   $U$  & 393 & & & \\
   $G^{10k}_{0.50}$ & p-value & \textbf{0.2814} & 0.0475 &     &      \\
            &   $U$  & 411 & 337 & & \\
    $G^{10k}_{0.25}$ & p-value & \textbf{0.2227} & 0.0377 &\textbf{ 0.4970} &    \\
            &   $U$  & 398 & 330 & 449 & \\
  $G^{10k}_{0.single}$ & p-value & \textbf{0.1604} &\textbf{ 0.4793} & 0.0150 & 0.0181 \\
            &   $U$  & 383 & 446 & 303 & 308 \\
\bottomrule
\end{tabular}
\end{table}

\begin{figure}[H]
\includegraphics[width=0.4\textwidth]{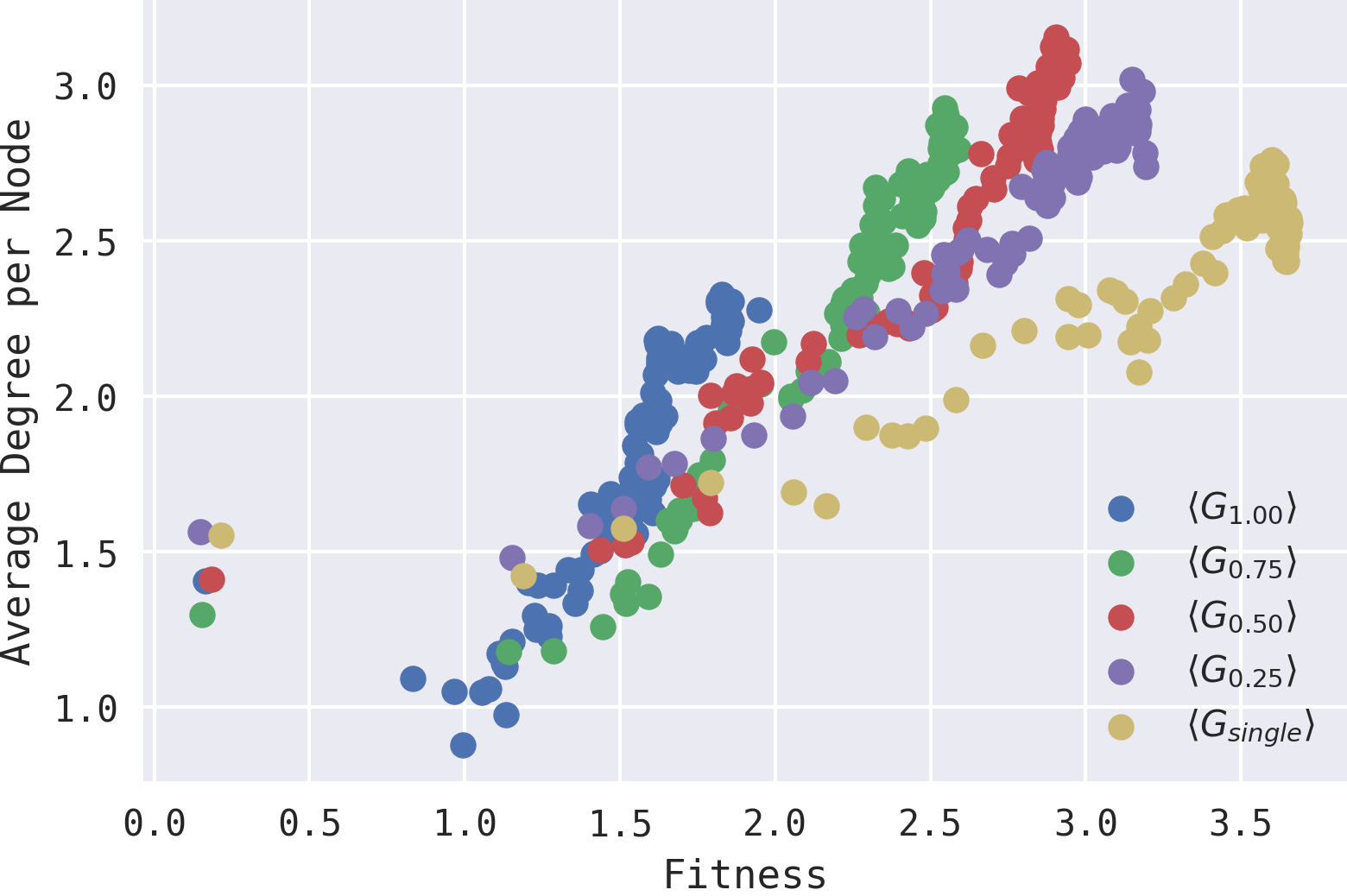}
\caption{ Fitness plotted against the average degree of all nodes in an animat's brain. One data point is the mean of the average degree over 30 experiments per generation evaluated at every $100^{th}$ generation).}
\label{fig:av_deg}
\end{figure}

\begin{table}[H]
\caption{Significance of the average degree of the animat's MB in the last generation:  p-values for the Mann-Whitney-Test testing the significance of the average degree per mechanism values between the groups ($n_{ij}=30$).}

\begin{tabular}{lrllll}
\toprule{}  &  & $G^{10k}_{1.00}$ & $G^{10k}_{0.75}$ & $G^{10k}_{0.50}$ & $G^{10k}_{0.25}$ \\

\midrule
    $G^{10k}_{0.75}$ &   p-value & \textbf{0.2954} &    &    &   \\
            &   $U$  & 437 & & & \\
   $G^{10k}_{0.50}$ & p-value & \textbf{0.2624} & \textbf{0.0789} &     &      \\
            &   $U$  & 403 & 378 & & \\
    $G^{10k}_{0.25}$ & p-value & \textbf{0.318}3 & \textbf{0.4612} & 0.0518 &    \\
            &   $U$  & 394 & 361 & 424 & \\
  $G^{10k}_{0.single}$ & p-value & 0.0038 & 0.0292 & 0.0001 & 0.0076 \\
            &   $U$  & 417 & 426 & 336 & 337 \\
\bottomrule
\end{tabular}
\end{table}

\begin{figure}[H]
\includegraphics[width=0.4\textwidth]{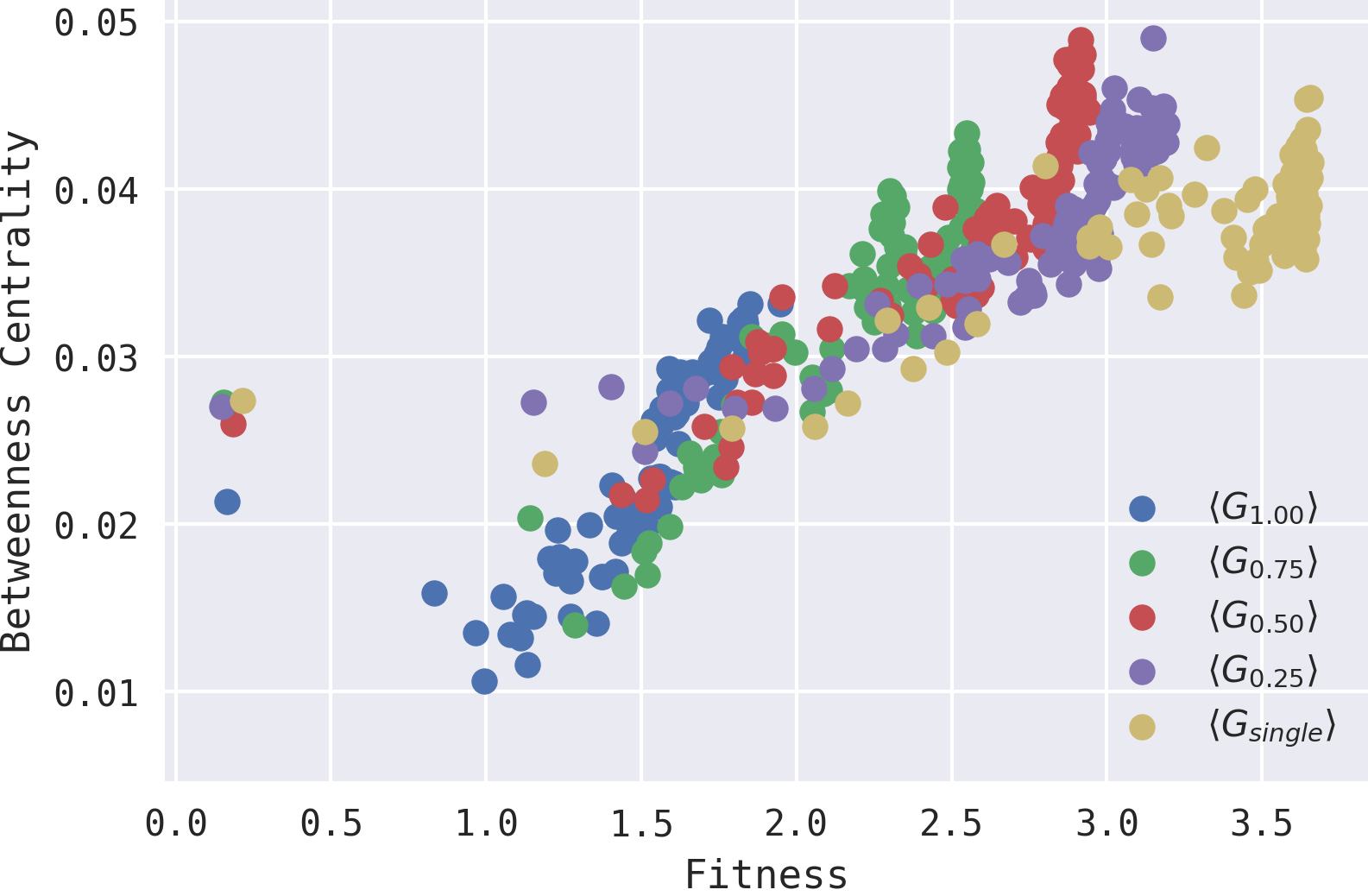}
\caption{ Fitness plotted against the average betweeness centrality of all nodes in an animat's brain. One data point is the mean of the average betweeness centrality over 30 experiments per generation evaluated at every $100^{th}$ generation).}
\label{fig:betw}
\end{figure}

\noindent

Since the Kruskal-Wallis test shows that there is no significant difference between the groups $G^{10k}_i$ ($H=4.0892, p=0.3941$) we do not provide the values of the Mann-Whitney U test.

\end{document}